\documentclass{article}

\usepackage[final,nonatbib]{neurips_2024}

\usepackage[utf8]{inputenc} 
\usepackage[T1]{fontenc}    
\usepackage{hyperref}       
\usepackage{url}            
\usepackage{booktabs}       
\usepackage{amsfonts}       
\usepackage{nicefrac}       
\usepackage{microtype}      
\usepackage{xcolor}         
\usepackage{graphicx}
\usepackage{placeins}

\usepackage[maxbibnames=5,minbibnames=3,maxcitenames=1,mincitenames=1,doi=false,url=false,eprint=false,natbib,sorting=none,giveninits]{biblatex}
\AtEveryBibitem{%
\ifentrytype{article}{
    \clearfield{issn}
    \clearfield{url}%
    \clearfield{urlyear}%
    \clearfield{month}%
    \clearfield{number}%
    \clearfield{language}%
}{}

\ifentrytype{inproceedings}{
    \clearfield{url}%
    \clearfield{urlyear}%
    \clearfield{month}
    \clearfield{address}%
    \clearfield{pages}%
    \clearfield{isbn}
}{}

\ifentrytype{misc}{
    \clearfield{primaryclass}%
    \clearfield{url}%
    \clearfield{issn}%
    \clearfield{urlyear}%
    \clearfield{month}
    \clearfield{journal}%
    \clearfield{doi}%
}{}
}

\DeclareSourcemap{
  \maps[datatype=bibtex]{
    \map{
      \step[fieldset=primaryclass,null]
    }
  }
}

\title{Evaluating the role of `Constitutions'\\for learning from AI feedback}

\author{%
  Saskia Redgate\\
 \vspace{-1.6ex} \\
 University of Oxford\\
{\small \texttt{saskia.redgate@gmail.com}}\\
  \And
  Andrew M. Bean\\
   \vspace{-1.6ex} \\
  University of Oxford \\
 {\small    \texttt{andrew.bean@oii.ox.ac.uk}}
  \And
  Adam Mahdi\\
   \vspace{-1.6ex} \\
  University of Oxford \\
{\small     \texttt{adam.mahdi@oii.ox.ac.uk}}
}

\begin{document}

\maketitle

\begin{abstract}
  The growing capabilities of large language models (LLMs) have led to their use as substitutes for human feedback for training and assessing other LLMs. These methods often rely on `constitutions', written guidelines which a critic model uses to provide feedback and improve generations. We investigate how the choice of constitution affects feedback quality by using four different constitutions to improve patient-centered communication in medical interviews. In pairwise comparisons conducted by 215 human raters, we found that detailed constitutions led to better results regarding emotive qualities. However, none of the constitutions outperformed the baseline in learning more practically-oriented skills related to information gathering and provision. Our findings indicate that while detailed constitutions should be prioritised, there are possible limitations to the effectiveness of AI feedback as a reward signal in certain areas.\\
  \raisebox{-0.3\height}{\includegraphics[width=0.35cm]{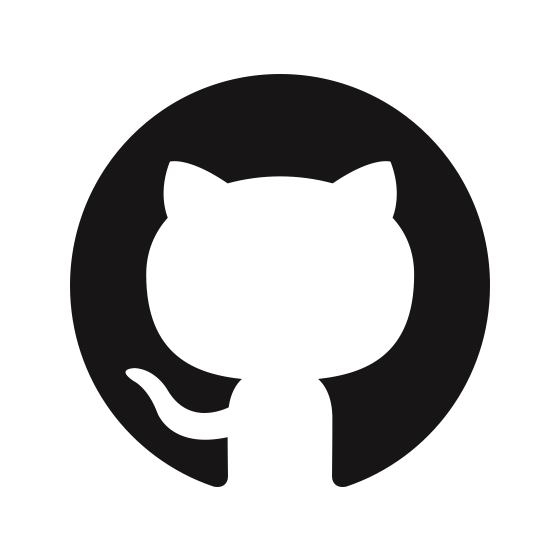}} \small \textbf{\mbox{Code:}} \href{https://github.com/saskia-rr/Evaluating-Constitutions}{github.com/saskia-rr/Evaluating-Constitutions}
\end{abstract}

\section{Introduction}
In current practice, pre-trained large language models (LLMs) are adapted with feedback learning to encode specific desirable abilities, especially conversational behaviours and safety alignment \cite{ouyang_training_2022, bai_constitutional_2022}. Learning from human feedback (e.g. RLHF) has been generally seen as the gold standard \cite{kirkPresentBetterFuture2023}, but this method can be prohibitively expensive, leading to the use of synthetic feedback paradigms such as `LLM as a Judge' \cite{chiangCanLargeLanguage2023} and `Constitutional AI' \cite{bai_constitutional_2022}. 

Using LLM-generated feedback involves asking a model to self-critique and generate revisions of previous work it has produced, typically based on a set of rules or ‘constitution’ \cite{bai_constitutional_2022}. Since these constitutions replace human interpretations of complex concepts and behaviours, it is important to consider how the content of the constitution impacts the results of the method. While previous work has shown that more specific constitutions are only marginally better than high-level goals in the case of broad values like `helpfulness/harmlessness' \cite{kundu_specific_2023}, we are additionally interested how well constitutions can shape specific socio-communicative behaviours. We draw on the case of medical practice, where principles for `patient-centered communication' \cite{king_best_2013} have been operationalised with detailed frameworks for the training and assessment of medical practitioners.

Medical uses of LLMs are an active area of study \cite{ayers_comparing_2023,hanSafeLargeLanguage2024, saabCapabilitiesGeminiModels2024, bean2023exploring, yang2024fine}, including the AIME model \cite{tu_towards_2024}, which incorporates an AI feedback learning approach to train social behaviours such as communication. We expand upon this work by exploring how different constitutions effect the ultimate quality of model generations. We compare four different test scenarios based on two different established clinical guidelines, broad role descriptions, and feedback in the absence of a constitution. We use iterative in-context learning to guide model generations based upon these constitutions, and then rate the quality of the final outputs in comparisons judged by humans. We find that using a more detailed constitution is more effective for improving patient-centered communication skills along emotive dimensions, but find no difference or worse performance along the more practically-oriented dimensions. With the increasing use of LLMs to replace human feedback and assessment, our results provide evidence for the effectiveness of these methods in many cases when used with detailed constitutions, but also indicate that AI feedback may be better suited to improving certain types of behaviours than others.

\section{Methods}
\subsection{In-context Learning with AI Feedback}

The core element of reinforcement learning with AI feedback is an iterative process of in-context learning, through which constitution-based feedback is used to create preferred model outputs \cite{bai_constitutional_2022, tu_towards_2024}. These outputs are subsequently used for fine-tuning, and the process can be repeated, but the primary impact of the constitutions takes place through this in-context learning. As such, we focus exclusively on the improvement of dialogues via in-context learning, with the expectation that better results in this portion of training would extend to better overall results. We use medical interviews as the foundation of these dialogues, based on two medical vignettes from the \textit{AgentClinic} dataset \cite{schmidgall_agentclinic_2024}.

\begin{figure}
    \centering
    \includegraphics[width=0.9\linewidth, trim=10pt 0pt 10pt 0pt, clip]{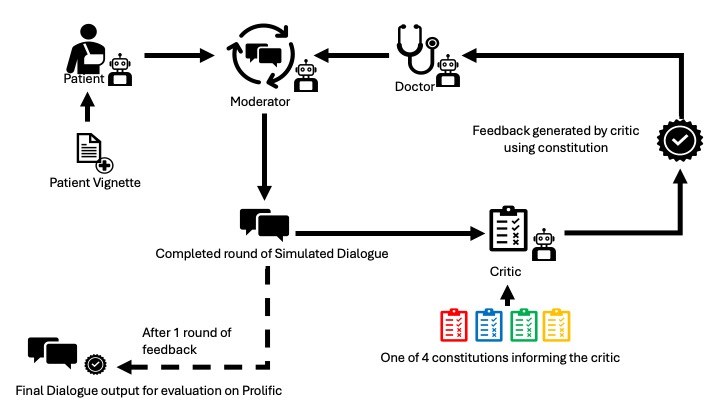}
    \caption{\textbf{Dialogue generation with in-context learning.} The \textit{Patient} model is given a vignette which is used to create a dialogue with a \textit{Doctor} model. A \textit{Moderator} model observes the conversation and intervenes when it sees a conversational indication that the interaction has ended. The conversation is then given to a \textit{Critic} model, which provides feedback based on one of the four different constitutions (Sec.~\ref{sec:constitutions}), and returns the feedback to the Doctor. This process is repeated for each vignette. The final conversations are collected and evaluated by 215 human raters recruited via \textit{Prolific} (Sec.~\ref{sec:human-eval}).}
    \label{fig:dialogue_generation}
\end{figure}

To perform in-context learning, we create an iterative loop using four different LLM agents, modelled after \citet{fu_improving_2023}, \citet{tu_towards_2024} and \citet{bai_constitutional_2022}. Each agent is an instance of Claude 3.5 Sonnet, queried via API. These roles (see Figure~\ref{fig:dialogue_generation}) include: \textit{Patient}, acting as a patient based on an AgentClinic vignette \cite{schmidgall_agentclinic_2024} in the system prompt which contains information about the symptoms and demographics of a patient `character'; \textit{Doctor}, who collects information and reaches a diagnosis for the `patient'; \textit{Moderator}, responsible for determining when the conversation between Doctor and Patient agent has ended; and \textit{Critic}, who provides feedback to the Doctor agent based on a chosen constitution. 

After the critic agent has given one round of feedback, and the patient and doctor have completed two conversations, we record the final conversation as the output to be assessed. We use this process to generate one complete conversation per constitution for each of the two vignettes. For fairness between constitutions, we excluded and replaced conversations where the patient model failed to follow the vignette by hallucinating symptoms or not acting as a patient. Complete prompt templates and parameters for each of the agents are included in Appendix~\ref{app:prompts} and vignettes are in Appendix~\ref{app:vignettes}.

\subsection{Constitutions}\label{sec:constitutions}
We compare four constitutions, as described below (for the full text, see Appendix~\ref{app:constitutions}). 1) \textit{Best Practices}, based on the widely-used `Patient-Centered Communication' framework established by \citet{king_best_2013}, is highly detailed and aligns with the criteria used to evaluate the final conversations. 2) \textit{Empathetic}, derived from EPITOME framework for empathetic text \cite{sharma_computational_2020}, is moderately detailed but focuses on only one aspect of socio-communicative skills. 3) \textit{Doctor}, inspired by \citet{kundu_specific_2023}, specifies only that the output should be in line with a good doctor, relying on the Critic for interpretation. 4) \textit{No Constitution} serves as a baseline, where the Critic provides feedback to improve the dialogue without specifying guidelines.

\subsection{Evaluation Framework}

The final conversations are compared according to the six categories of the `Patient-Centered Communication' framework \cite{king_best_2013}, with the relevant questions adapted from \citet{reeve_psychometric_2017} and \citet{moser_patient-centered_2022}. The categories are shown in Table~\ref{tab:pcc}.

\begin{table}[t!]
    \centering
    \renewcommand{\arraystretch}{1.3}
    \begin{tabular}{p{0.3\textwidth}p{0.65\textwidth}}
        \hline
        \textbf{Dimension} & \textbf{Question}\\ \hline 
         Fostering the Relationship & Had open and honest communication with the patient 
         \\ 
         Gathering Information & Give the patient the chance to ask all the health-related questions they had \\
         Providing Information & Explain things to the patient in a way they could understand \\
         Decision Making & 
         Involve the patient in decisions about their health care as much as they wanted \\
         Enabling disease and treat\-ment-related behaviour & Made sure the patient understood the things they needed to do to take care of their health \\ 
          Responding to emotions & Give the attention the patient needed to their feelings and emotions \\
          \hline
    \end{tabular}
    \smallskip
    \caption{\textbf{Dimensions of Patient-Centered Communication.} Each dimension of Patient-Centered Communication (PCC) Best Practices \cite{king_best_2013} and the corresponding evaluation question.}
    \label{tab:pcc}
\end{table}

For each dimension, we collect pairwise ratings between the generated conversations. We then use a Bradley-Terry model \cite{chiangChatbotArenaOpen2024} to estimate an underlying parameter of the quality of the conversations along each dimension.

\subsection{Human Evaluation}
\label{sec:human-eval}

To evaluate the final conversations, we recruited 215 human raters from Prolific. Each participant is presented with two randomly selected conversations based on different constitutions, and asked to make  comparisons between them according to the `Patient-Centered Communication' framework as well as providing a holistic preference. Participants repeat this twice, seeing one conversation for each constitution, but not all six possible pairings. Participants are paid £2.75 for an average of 13 minutes of time. Due to the length of the conversations being compared, we required participants to answer one comprehension check question per conversation, and we excluded 2 participants who failed more than once. We did not exclude participants who skipped other questions, leading to slight imbalances between the number of ratings per question and pair. We excluded 16 participants who started but did not complete the survey, an attrition rate of 7\%. This research was pre-approved and carried out in line with institutional ethics approval (reference number OII\_C1A\_24\_203).

\section{Results}

\begin{figure}
    \centering
    \includegraphics[width=\linewidth]{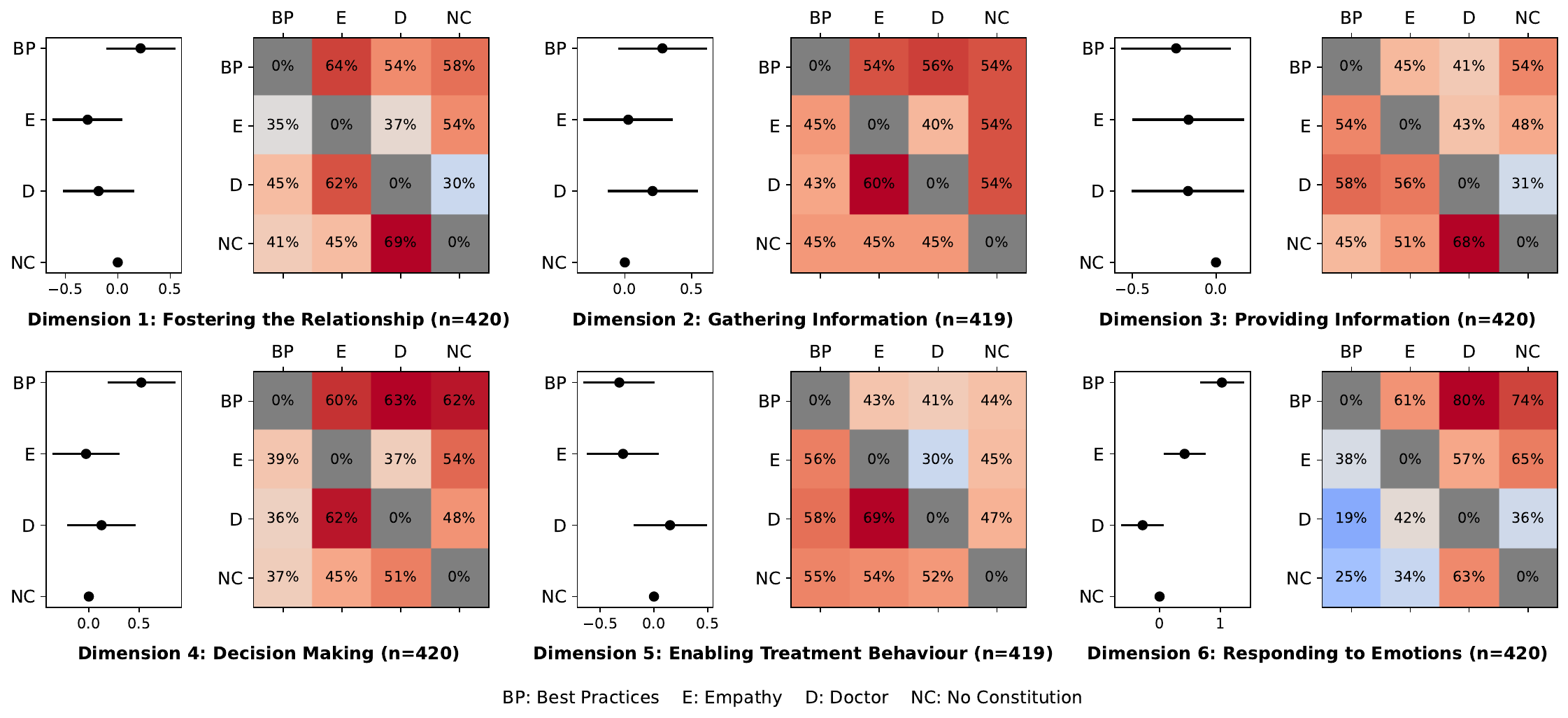}
    \caption{\textbf{Preferred Constitutions.} In each subplot, we show the percentage of respondents preferring each conversation as a heatmap, alongside the estimated values from a Bradley-Terry model. We set the `No Constitution' group as a reference point. Error bars represent a 95\% confidence interval, not adjusted for multiple comparisons.}
    \label{fig:results}
\end{figure}

In Figure~\ref{fig:results}, we show the rate at which the conversations generated according to each constitution are preferred to the others for each dimension of evaluation, alongside the estimated parameters for a Bradley-Terry model.

The Best Practices constitution is preferred to the other constitutions for `Fostering the Relationship', `Decision Making', and `Responding to Emotions'. There is not a clear difference between the constitutions for `Gathering' and `Providing Information'. For `Enabling treatment behaviour', the Best Practices constitution leads to worse results than the non-specific doctor constitution and the empty constitution. When holistically selecting a most-preferred constitution, there was not a clear pattern across participants, though many participants indicated a dislike of verbose or overly emotive responses even when rating them as more `empathetic'.

\section{Discussion}\label{sec:discussion}
For the emotionally-oriented dimensions (Fostering the Relationship, Decision Making, and Responding to Emotions) of patient-centered communication, we found that the most specific constitution led to the most human-preferred dialogues. This is consistent with previous work comparing constitutions in the case of `harmlessness' \cite{kundu_specific_2023}, and indicates that efforts to create detailed constitutions are likely to improve the outcomes of AI feedback methods. This is also supported by the poor performance of the generic ``Doctor'' constitution, which is indistinguishable from the ``No constitution'' treatment in all six dimensions, and the success of the `Empathy' constitution in `Responding to Emotions', but not the other categories. 

We do not see the same improvements for the more practically-oriented dimensions, where the LLM needs to manage information exchange with the patient. These type of behaviours may be more difficult for language models to judge and learn, as they involve planning and theory of mind, while emotional signals may be imitated by adding sensitive-sounding phrases \cite{ayers_comparing_2023}.

We also note that qualitative feedback from participants who did not like the verbosity of the models reveals that aspects of the reward function such as sentence length may be intuitive to humans but not LLMs \cite{tu_towards_2024}, and that human preferences remain difficult to measure well.


In this study, we focused on comparing four specific constitutions in the case of patient-centered communication in medicine. For each constitution, we used only two dialogues, limiting the generalisability. This is partially compensated by having six different axes of comparison, showing that the dialogues are improved in several, but not all cases. While we argue that in-context learning is the key mechanism for RLAIF, fine-tuning based on a collection of examples would allow a model to learn behaviours which are not present in every example. As such, a wide range of small improvements may be aggregated to achieve better results than what we observe in a single interaction with in-context learning.

\printbibliography

\appendix

\newpage
\section{Constitutions}
\label{app:constitutions}

We provide the full text of the four constitutions.

\begin{figure}[ht!]
    \centering
    \framebox{
    \begin{minipage}{\textwidth}
    \textbf{Constitution 1: Best Practices}

    You are an AI critic providing feedback to a doctor after they have completed a conversation with a patient. All of the previous conversation is the completed conversation. Provide feedback based on the following guidelines:

\subsubsection*{1. Fostering the relationship}
\paragraph{Roles:}
\begin{itemize}
  \item Build rapport and connection
  \item Appear open and honest
  \item Discuss mutual roles and responsibilities
  \item Respect patient statements, privacy, and autonomy
  \item Engage in partnership building
  \item Express caring and commitment
  \item Acknowledge and express sorrow for mistakes
\end{itemize}

\paragraph{Skills:}
\begin{itemize}
  \item Greet patient appropriately
  \item Use appropriate language
  \item Encourage patient participation
  \item Show interest in the patient as a person
\end{itemize}

\subsubsection*{2. Gathering information}
\paragraph{Roles:}
\begin{itemize}
  \item Attempt to understand the patient's needs for the encounter
  \item Elicit full description of major reason for visit from biologic and psychosocial perspectives
  \item Elicit patient's full set of concerns
  \item Elicit patient's perspective on the problem/illness
  \item Explore full effect of the illness
\end{itemize}

\paragraph{Skills:}
\begin{itemize}
  \item Ask open-ended questions
  \item Allow patient to complete responses
  \item Listen actively
  \item Clarify and summarize information
  \item Inquire about additional concerns
\end{itemize}

\subsubsection*{3. Providing information}
\paragraph{Roles:}
\begin{itemize}
  \item Seek to understand patient's informational needs
  \item Share information
  \item Overcome barriers to patient understanding (language, health literacy, hearing, numeracy)
  \item Facilitate understanding
  \item Provide information resources and help patient evaluate and use them
\end{itemize}

  \end{minipage}
    }
\end{figure}

\begin{figure}[ht!]
    \centering
    \framebox{
    \begin{minipage}{\textwidth}
    \textbf{Constitution 1: Best Practices (continued)}

\paragraph{Skills:}
\begin{itemize}
  \item Explain nature of problem and approach to diagnosis and treatment
  \item Give uncomplicated explanations and instructions
  \item Avoid jargon and complexity
  \item Encourage questions and check understanding
  \item Emphasize key messages
\end{itemize}

\subsubsection*{4. Decision making}
\paragraph{Roles:}
\begin{itemize}
  \item Prepare patient for deliberation and enable decision making
  \item Outline collaborative action plan
  \item Encourage patient to participate in decision making
\end{itemize}

\paragraph{Skills:}
\begin{itemize}
  \item Outline choices
  \item Explore patient's preferences and understanding
  \item Reach agreement
  \item Identify and enlist resources and support
  \item Discuss follow-up and plan for unexpected outcomes
\end{itemize}

\subsubsection*{5. Enabling disease- and treatment-related behavior}
\paragraph{Roles:}
\begin{itemize}
  \item Assess patient's interest in and capacity for self-management
  \item Provide advice (information needs, coping skills, strategies for success)
  \item Agree on next steps
  \item Assist patient to optimize autonomy and self-management of his or her problem
  \item Arrange for needed support
  \item Advocate for and assist patient with health system
\end{itemize}

\paragraph{Skills:}
\begin{itemize}
  \item Assess patient's readiness to change health behaviors
  \item Elicit patient's goals, ideas, and decisions
\end{itemize}

\subsubsection*{6. Responding to emotions}
\paragraph{Roles:}
\begin{itemize}
  \item Facilitate patient expression of emotional consequences of illness
  \item Acknowledge and explore emotions
  \item Express empathy, sympathy, and reassurance
  \item Provide help in dealing with emotions
\end{itemize}

\paragraph{Skills:}
\begin{itemize}
  \item Assess psychological distress
\end{itemize}

Your feedback will only be seen by the doctor. Give your feedback in 100 words or less.
\end{minipage}
    }
\end{figure}

\begin{figure}[ht!]
    \centering
    \framebox{
    \begin{minipage}{\textwidth}
    \textbf{Constitution 2: Empathetic}

You are an AI critic providing feedback to a peer. All of the previous conversation is the completed conversation. Provide feedback based on the following guidelines:

\begin{itemize}
  \item \textbf{Emotional Reactions:} Expressing emotions such as warmth, compassion, and concern, experienced by peer supporter after reading seeker's post
  \item \textbf{Interpretations:} Communicating an understanding of feelings and experiences inferred from seeker's response
  \item \textbf{Explorations:} Improving understanding of the seeker by exploring the feelings and experiences not stated in the reply. Showing an active interest in what the seeker is experiencing and feeling and probing gently
\end{itemize}

Your feedback will only be seen by the peer. Give your feedback in 100 words or less.

    \end{minipage}
    }
\end{figure}

\begin{figure}[t!]
    \centering
    \framebox{
    \begin{minipage}{\textwidth}
    \textbf{Constitution 3: Doctor}
    
You are an AI critic providing feedback to a doctor after they have completed a conversation with a patient. Give feedback as you see fit. Your feedback will only be seen by the doctor. Give your feedback in 100 words or less.

    \end{minipage}
    }
\end{figure}

\begin{figure}[t!]
    \centering
    \framebox{
    \begin{minipage}{\textwidth}
    \textbf{Constitution 4: No Constitution}
    
You are an AI critic providing feedback. Give feedback as you see fit. Give your feedback in 100 words or less.
    \end{minipage}
    }
\end{figure}
\FloatBarrier
\section{Model Agent Specifications}
\label{app:prompts}

All models used in this study were copies of the Claude 3.5 Sonnet model, accessed via API. Temperature was set to 1.0. The total inference cost was \$60.

We prompted each of the agents using the following templates, based on \citep{tu_towards_2024}. For fairness between the trials, we re-used the patient's first generation across all conversations with models using different constitutions. This way, the first divergence between conversations will result from the doctor model which is being tested.

\begin{figure}[ht!]
    \centering
    \framebox{
    \begin{minipage}{\textwidth}

\textbf{Patient}

You are a patient chatting with a doctor over an online chat interface. The doctor has never met you before. This is your profile:\\

Demographics: \textit{...}

Overview: \textit{...}

Primary Symptoms: \textit{...}

Secondary Symptoms: \textit{...}

Medical History: \textit{...}

Social History: \textit{...}

Key Review of Vitals: \textit{...}\\

Using the profile you should answer as the patient. Do not reveal you are an AI chatbot. Give your responses in 60 words or less.
\end{minipage}
}
\end{figure}

\begin{figure}[ht!]
    \centering
    \framebox{
    \begin{minipage}{\textwidth}

\textbf{Doctor}

You are a Doctor speaking to a patient over an online chat interface. You know nothing about the patient in advance. Respond in single-turn responses to understand their symptoms and find a diagnosis. You should provide a diagnosis to the patient. You are the healthcare provider. Do not tell the patient to see a healthcare provider. Do not reveal you are an AI bot. Give your responses in 50 words or less.
\end{minipage}
}
\end{figure}

\begin{figure}[ht!]
    \centering
    \framebox{
    \begin{minipage}{\textwidth}
        \textbf{Moderator}

        You are a helpful AI agent which is monitoring a simulated conversation between a Doctor and a Patient. You should stop the conversation when you feel a natural conclusion has been reached. Do not terminate the conversation if there are any open questions left unanswered.
    \end{minipage}
}
\end{figure}

\begin{figure}[ht!]
    \centering
    \framebox{
    \begin{minipage}{\textwidth}
        \textbf{Critic}

You are an AI critic providing feedback \{INSERT CONSTITUTION\} Give your feedback in 100 words or less.

\end{minipage}
}
\end{figure}

To make sure the feedback has been acknowledged by the Doctor, additional leading text is added to the critic feedback and the next turns in conversation are forced such that in the Doctors context the feedback has been received. This is following the prompt structure on receiving feedback based on \cite{fu_improving_2023}.

\noindent\fbox{%
\parbox{\textwidth}{%
    \textbf{USER} ``Here is feedback on your previous interaction with the patient: \{CRITIC FEEDBACK\} Incorporate this feedback into your responses in the next turn of conversation''\\
    \textbf{ASSISTANT (DOCTOR)} ``I understand and have acknowledged the feedback. I will incorporate it into the next turn of the conversation.''\\
    \textbf{USER} ``The next round of conversation is about to start.''\\
    \textbf{ASSISTANT (DOCTOR)} ``Hello, how can I help you today?''
}
}
\section{Platform Screenshots}
\label{app:platform}

\begin{figure}[t!]
    \centering
    \includegraphics[width=\linewidth]{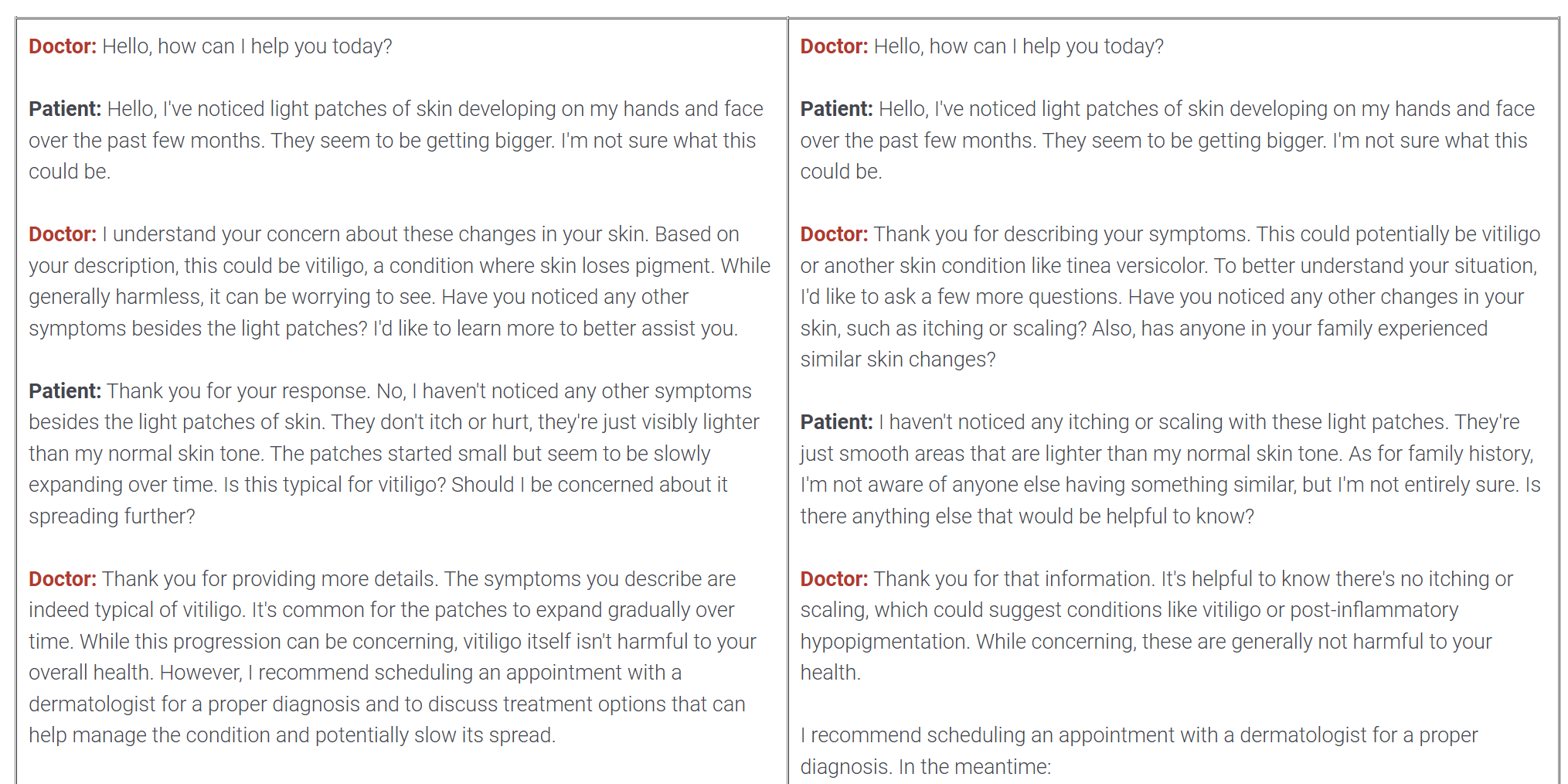}
    \caption{\textbf{Side-by-side dialogues.} The pairs of dialogues to be compared are presented side by side. The doctor is highlighted in red for visual clarity.}
    \label{fig:dialogues}
\end{figure}

Below are screenshots of the experimental platform where human feedback was collected. Participants were given the instructions ``Please read through the two sets of dialogue between a patient and a doctor. After reading, please answer the questions below.'' The dialogues were shown side-by-side as seen in Figure~\ref{fig:dialogues}.

After reading the dialogues, participants were further instructed: ``Once you have finished reading the two sets of dialogue, please answer the questions below. You will be asked 4 questions in total. 

The first 2 questions will check you have read both pieces of dialogue thoroughly.

The next 2 will ask for your opinion of the Doctor in the 2 pieces of text. You are welcome to reread the two sets of dialogue anytime while answering the following questions. ''

\begin{figure}[ht!]
    \centering
    \includegraphics[width=0.9\linewidth]{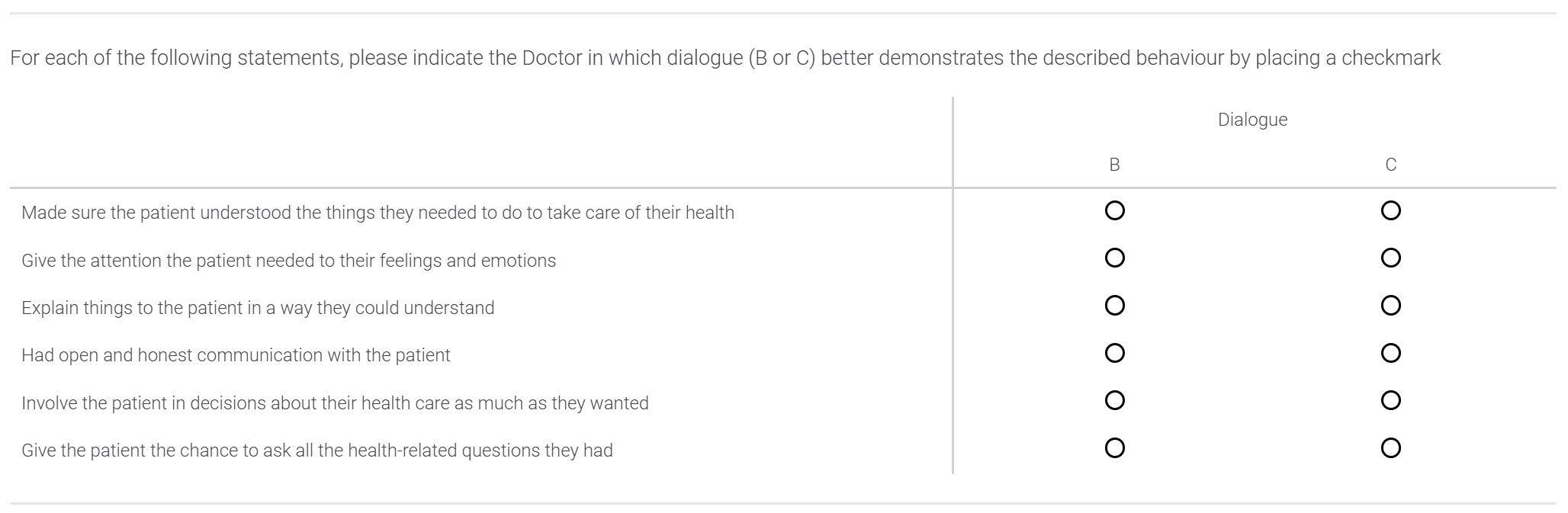}
    \caption{\textbf{Preference ratings.} Participants choose which dialogue they preferred for each aspect of the patient-centered communication framework.}
    \label{fig:responses}
\end{figure}

The comprehension checks are given as multiple choice questions based on the content of the passage. Participants then complete forced-choice comparisons between the two dialogues for each aspect of the patient-centered communication framework as shown in Figure~\ref{fig:responses}.

\section{Vignettes}
\label{app:vignettes}

We used two vignettes in this study, taken from the \textit{AgentClinic} dataset \cite{schmidgall_agentclinic_2024}. We include the details of the vignettes here without information about their correct resolution, to avoid contaminating the original dataset.

\textbf{Vignette 1:}

\textit{Demographics:} 19-year-old Caucasian male\\
\textit{Overview :} The patient reports noticing gradually developing patches of lighter skin on his hands and face over the past few months. These patches seem to be expanding in size. He denies any pain, itching, or other discomfort in the areas. No recent illnesses, medication changes, or significant sunburns.\\
\textit{Primary Symptoms:} Hypopigmented skin patches\\ 
\textit{Secondary Symptoms:} No discomfort in the affected areas, Gradual increase in size of the patches\\ 
\textit{Medical History:} No significant past medical history. The patient is otherwise healthy with no chronic conditions. \\
\textit{Social History:} Full-time university student, non-smoker, and occasional alcohol use.\\
\textit{Key Review of Vitals:} Denies recent flu-like symptoms, fever, weight loss, changes in vision, hair loss, or history of skin cancer in the family.

\textbf{Vignette 2:}

\textit{Demographics:} 45-year-old female\\
\textit{Overview:} The patient reports a 2-week history of rectal bleeding occurring daily with bowel movements. She denies any pain with defecation and does not present with any other complaints.\\
\textit{Primary Symptoms:} Rectal bleeding daily with bowel movements\\
\textit{Secondary Symptoms:} No pain with defecation
\textit{Medical History:} The patient’s past medical history is unremarkable except for 5 normal vaginal deliveries. \\
\textit{Social History:} Information not specified. \\
\textit{Key Review of Vitals:} The patient denies any changes in bowel habits, abdominal pain, weight loss, or other systemic symptoms.

\end{document}